\documentclass{singlecol-new}
\usepackage{amsmath}
\usepackage{algorithm}
\usepackage{algpseudocode}
\usepackage{graphicx}
\usepackage[numbers]{natbib}
\usepackage{mathrsfs}
\usepackage[colorinlistoftodos]{todonotes}
\usepackage{placeins}

\theoremstyle{TH}{

}

\theoremstyle{THrm}{

}

\theoremstyle{THhit}{

}

\makeatletter
\def\theequation{\arabic{equation}}

\makeatother

\begin{document}%

\setcounter{page}{1}

\LRH{P. Jackovich et~al.}

\RRH{Comparing Greedy Subtour Elimination Methods for TSP}







\title{Comparing Greedy Constructive Heuristic Subtour Elimination Methods for the Traveling Salesman Problem}

\authorA{Petar D. Jackovich}
\affA{Department of Operational Sciences,\\ Air Force Institute of Technology,\\ Wright-Patterson AFB, OH, USA \\
E-mail: petar.jackovich@AFIT.edu}

\authorB{Bruce A. Cox}
\affB{Department of Operational Sciences,\\ Air Force Institute of Technology,\\ Wright-Patterson AFB, OH, USA \\
E-mail: Bruce.Cox@AFIT.edu}

\authorC{Raymond R. Hill}
\affC{Department of Operational Sciences,\\ Air Force Institute of Technology,\\ Wright-Patterson AFB, OH, USA \\
E-mail: rayrhill@gmail.com}

\begin{abstract}
This paper further defines the class of fragment constructive heuristics used to compute feasible solutions for the Traveling Salesman Problem into arc-greedy and node-greedy subclasses. Since these subclasses of heuristics can create subtours, two known methodologies for subtour elimination on symmetric instances are reviewed and are expanded to cover asymmetric problem instances. This paper introduces a third novel methodology, the Greedy Tracker, and compares it to both known methodologies. Computational results are generated across multiple symmetric and asymmetric instances. The results demonstrate the Greedy Tracker is the fastest method for preventing subtours for instances below 400 nodes.  A distinction between fragment constructive heuristics and the subtour elimination methodology used to ensure the feasibility of resulting solutions enables the introduction of a new node-greedy fragment heuristic called Ordered Greedy. 
\end{abstract}

\KEYWORD{Traveling Salesman Problem; Heuristic; Fragment Constructive Heuristic; Multiple Fragment Heuristic; Constructive Heuristic; Subtour Elimination; Exhaustive Loop; Arc-Greedy; Node-Greedy; Greedy Tracker; Ordered Greedy; }


%
%

\maketitle

\section{Introduction}
\par Applegate et al \citep{applegate2006traveling} describes the traveling salesman problem as, ``Given a set of cites along with the cost of travel between each pair of them, the traveling salesman problem, or TSP for short, is the problem of finding the cheapest way of visiting all the cities and returning to the starting point." It can also be mathematically defined as, given a complete graph $G = (V,E)$, cities are represented via the graph vertices $g \in G$ , and edges $e \in E$ represent the paths between the cities where the edge weights are the distances between each city: What is the shortest tour that visits all vertices once and returns to the starting vertex? When traditional linear programming methods were applied to the TSP, intractability issues arose \citep{applegate2006traveling}. It has since been shown this is because the TSP falls into a class of known computationally `hard' problems called NP-Complete \citep{rego2011traveling}.  No one has yet developed an efficient method for universally solving large instances of NP-complete problems to optimality \citep{khan2012multilevel}. The inclusion of the TSP in the set of NP-complete problems motivates the usage of other solving techniques such as heuristics.

\par  One class of heuristics utilize a greedy-type methodology, where the best choice, according to a predefined parameter, is selected at each step of the method. An example of a greedy-type method for the TSP is the Multiple-Fragment heuristic \citep{bentley1992fast}, where the shortest available arc is iteratively added to form a  tour. However, this greedy heuristic runs the risk of generating subtours, or disconnected tours of less than size N (where N is the number of nodes present in the graph) that prevent a single continuous tour from being formed. Some research has been completed to develop methodologies that avoid subtours when utilizing the arc-greedy heuristic \citep{bentley1992fast, wang2018distance}.

\par This paper further defines the class of Fragment Heuristics defined by Bentley \citep{bentley1992fast} and discusses the importance of subtour elimination methodologies within this class. A novel subtour elimination methodology for the arc-greedy heuristic is introduced and compared to two known subtour elimination methodologies. Extensions to all three subtour elimination methodologies to handle asymmetric instances are introduced. Computational results are generated across multiple TSP instances for each method to compare run times by size and geometry.

\section{Literature Review}
\subsection{Types of TSP Construction Heuristics}
Bentley's \citep{bentley1992fast} 1990s paper, in addition to many other contributions, established classifications for TSP constructive heuristics. Bentley breaks TSP construction heuristics into three general categories: heuristics that grow fragments, heuristics that grow tours, and heuristics based on trees. The latter two classes each ensure viable TSP tours utilizing a consistent methodology for each heuristic in its class. For example, the class "heuristics that grow tours" contains heuristics that follow the Insertion or Addition expansion rules. These heuristics each start with a partial tour consisting of a single node which is then gradually grown by adding new nodes. This is accomplished by deleting an edge and adding two new edges, according to the specified rule-set, to connect the new node to the current tour, ensuring a valid TSP tour is constructed. However, the "heuristics that grow fragments" class does not have a consistent methodology to avoid premature partial circuits, or subtours.

\subsection{Heuristics that Grow Fragments}
In this section we will briefly introduce the heuristics described in the ``Heuristics that Grow Fragments" section of Bentley's \citep{bentley1992fast} paper.

\subsubsection{Nearest Neighbor}
\par The nearest neighbor (NN) heuristic was first applied to the TSP in a 1954 paper by Flood \citep{flood1956traveling} but was introduced as the "next closest city method." The process was later refined by Dacey \citep{dacey1960letter} and coined with its eventual name. The NN starts at an arbitrary city, and successively visits the closest unvisited city. Note that the nearest neighbor heuristic maintains a single path fragment that originates at the predetermined starting city, and cannot be closed into a cycle until every node has been visited. Therefore the decision of ``which arc to add" is limited to only those arcs that leave the current end node of the fragment, this yields an algorithm run time of \textit{O}($N^2$). Future work by Bentley \citep{bentley1992fast} allowed this heuristic to perform in \textit{O}(\textit{N} log \textit{N}). This methodology allows NN to quickly create an initial tour which avoids sub tours. However, NN is extremely sensitive to the choice of starting node, especially in larger instances. Bentley also introduces a second variant of NN called the Double-Ended Nearest Neighbor (DENN) which allows the fragment to grow from both ends.

\subsubsection{Multiple-Fragment}
\par The Multiple-Fragment heuristic (MF) was first introduced by Papadimitiou and Steiglitz \citep{papadimitriou1982combinatorial} as a modification of a process first seen in a 1968 paper by Steiglitz and Weiner \citep{steiglitz1968some}. The heuristic is a more complex greedy-type TSP heuristic where all edges of the graph are sorted from shortest to longest. Edges are then added to the tour starting with the shortest arc as long as the addition of this arc will not make it impossible to complete a tour. Specifically, this means avoiding adding edges that make early cycles, and also avoiding creation of vertices of degree three. This process, as originally proposed, required \textit{O}(\textit{$N^2$} log \textit{N}) time. However, Bentley was able to speed up this process to \textit{O}(\textit{N} log \textit{N}) \citep{bentley1992fast} in a paper introducing his MF version. This yields a similar run time to NN while maintaining a similar worst case solution quality. MF's tour construction methodology causes the heuristic to only produce a single solution for each instance where NN can arrive at different solutions based on a different starting point. When compared to the average NN solution over all starting points, MF tends to outperform NN on an instance-to-instance basis \citep{bentley1992fast, okano1999new}.

\section{Methodology/Discussion} 
At present it is entirely unclear when referencing the "Greedy Heuristic" for the TSP if the heuristic under discussion is NN or MF. This is acknowledged in \citep{aarts2003local}.  Much of this confusions seems to draw from the poor naming conventions used with respect to how a greedy methodology works. In accordance with the framework established by Talibi \citep{talbi2009metaheuristics} constructive heuristics involve two choices. First, determine a set of elements, $S_j = \{e_{1,j},e_{2,j},...,e_{p,j}\}$ which comprise the neighborhood of the current solution, that is the set of possible choices at each iteration $j$. Second, define a methodology to choose an element $e_{i,j}$ from this set $S_j$. Thus the framework of how this set $S_j$ is defined determines in large part how the heuristic constructs the solution. In the case of these TSP fragment heuristics we propose a more comprehensive classification that splits this class into two greedy-type methodologies; node-greedy and arc-greedy heuristics based on how the set $S_j$ is defined.

\subsection{Node-Greedy}
For the first proposed fragment greedy type, node-greedy, the set of arcs available at each iteration is defined by a given node.  The key defining characteristic of this sub-class is that the set $S_j$ is limited to the arcs incident to a given node. Depending on the measure of merit being greedily optimized this sub-class results in different heuristics. NN is an example of this node-greedy sub-class, where the starting node is user defined, from that point on the elements of $S_j$ are defined as the arcs incident to the node at the head of the fragment. The choice of element from $S_j$ associated with NN is the shortest arc.

\subsection{Arc-Greedy}
\begin{figure}[h!]
	\centering
    \includegraphics[width=12cm]{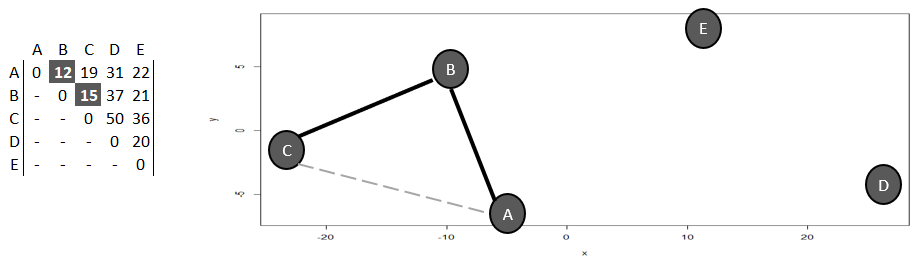}
    \caption{Multiple-Fragment arc-greedy heuristic's can generate subtours}
    \label{sub1}
\end{figure}

The arc-greedy type of fragment greedy TSP heuristic considers the set of all available arcs at each iteration, which will not make it impossible to complete a tour. The first criteria to narrow the scope of this set is to identify all arcs that will not cause a node to have a degree of more than 2 when added. The second criteria, which is much more difficult, is to identify what arcs will cause a tour of less than size N, or subtour, to form. Consider the following tour construction utilizing the multiple-fragment arc-greedy constructive heuristic methodology on a 5 node TSP instance with only the degree criteria. After adding the first two shortest arcs A-B and B-C, we can see from the distance matrix that arc A-C is the next shortest and still ensures that all nodes in the graph do not exceed a degree of 2. However, adding this arc creates a subtour, which would prevent the heuristic from ever constructing a feasible TSP tour (see Figure \ref{sub1}).
Part of the confusion with regards to the ``Greedy Heuristic" revolves around Bentley's MF being viewed as what we describe as \emph{the} arc-greedy heuristic.  However, we argue the true significance of Bentley's MF is the development of a subtour tracking and elimination methodology, which can then be applied to any member of the fragment class of TSP constructive heuristics.

\subsection{Subtour Tracking and Elimination Methodologies}
In addition to Bentley's MF methodology \citep{bentley1992fast}, a well known, though to the best of our knowledge, undocumented methodology we dub exhaustive loop also can be used to track and eliminate subtours. This paper introduces a third novel method for eliminating subtours while using an arc-greedy constructive heuristic we call the Greedy Tracker (GT). 

\subsubsection{Directional vs. Non-Directional}
\par  TSP instances can be either symmetric or asymmetric, similarly subtour tracking methodologies can be either directional or non-directional. Non-directional subtour tracking methodologies construct a tour with no regard to the direction of travel for each arc while ensuring no node has a degree of more than two. This methodology can only be used with symmetric TSP instances where the distance to travel from node to node is equal in both directions. This poses some computational advantages as only $n*(n+1)/2$ arcs need to be initially sorted. Directional subtour tracking methodologies can be used on either symmetric or asymmetric instances when the direction of arc travel is either of importance to the final solution and/or takes different costs to travel based on direction. In a directional scenario, all arcs of each direction $n^2-n$, are sorted from shortest to longest and rather than tracking the total degree of each node, each node can only be entered and left once, ensuring a continuous direction throughout the tour.

\subsubsection{Multiple-Fragment}
The Non-Directional variant of MF is well documented throughout TSP literature \citep{bentley1992fast, wang2018distance, okano1999new}. However, a formalization of the Directional variant is missing from this literature. This process appears to have been utilized in a 1999 paper by Glover et al \citep{glover2001construction}. However, neither pseudocode nor an explicit description of altering the methodology for directional instances is available. The primary alterations are to consider all arcs of each direction and the Degree array is split into `To' and `From' arrays. The process then continues as described by Bentley \citep{bentley1992fast}, where after each arc is added the tails of the associated fragment are updated to ensure no subtours are formed. Pseudocode for this modified methodology is included as Algorithm \ref{fig:MFPseudo}.

\begin{algorithm}
\caption{Multi-Fragment (Directional) Pseudocode}
\label{fig:MFPseudo}
\begin{algorithmic}[1]
\State Initialize Variables
\State Sort edges(i,j): Shortest to Longest
\While {Nodes.Visited \textless Size-1}
\If {From[i] = 0 \& To[j] = 0 \& Tail[i] != j}	
\State		Add edge(i,j)
\If {From[i]+To[i]=0 \& From[j]+To[j]=0}
\State			tempTaili = Tail[i]
\State			tempTailj = Tail[j]
\State			Tail[i] = tempTailj
\State		Tail[j] = tempTaili
\ElsIf {From[i]+To[i]=1 \& From[j]+To[j]=0}
\State			tempTaili = Tail[i]
\State		Tail[tempTaili] = Tail[j]
\State			Tail[j] = tempTaili
\State		Tail[i] = 0
\ElsIf {From[i]+To[i]=0 \& From[j]+To[j]=1}
\State		tempTailj = Tail[j]
\State			Tail[tempTailj] = Tail[i]
\State		Tail[i] = tempTailj
\State		Tail[j] = 0
\ElsIf {From[i]+To[i]=1 \& From[j]+To[j]=1}
\State			tempTaili = Tail[i]
\State		tempTailj = Tail[j]
\State		Tail[tempTaili] = tempTailj
\State	Tail[tempTaili] = tempTailj
\State	Tail[i] = 0
\State	Tail[j] = 0
\EndIf
\State	From[i] = From[i] + 1
\State	To[j] = To[j] + 1
\State Nodes.Visited = Nodes.Visited + 1
\EndIf
\State	Next Edge in List
\EndWhile
\State Connect Hamilton Path
\end{algorithmic}
\end{algorithm}

\subsubsection{Exhaustive Loop}
The Exhaustive Loop (EL) is not well documented in academic literature, often simply referenced as "the standard way." To the best of our knowledge no formal coverage of this method exists in literature. EL  cycles through every edge in the fragment containing the most recently added edge. Once an edge $e_{ij}$ is added to the partial tour, node \textit{i} is identified as the "start node" and node \textit{j} will be set as "current node." A trace along the current partial tour then begins. At each step of the trace the "current node," node \textit{j}, is checked to see if it is connected to another node \textit{k} via edge $e_{jk}$ in the partial tour. If it is, then node \textit{k} becomes the new "current node." If the trace returns back to the "start node" in under \textit{N} steps, where \textit{N = |V|} (the number of nodes in the instance), then the added edge $e_{ij}$ has created a subtour and is an illegal edge. If no edge leaves the "current node" the addition of edge $e_{ij}$ is valid and the current portion of the tour is still a fragment. Each time an edge is added, a count is incremented and the process continues until N-1 edges have been added upon which the last two endpoints are connected.
\par
\begin{figure}[h!]
	\centering
    \includegraphics[width=12.5cm]{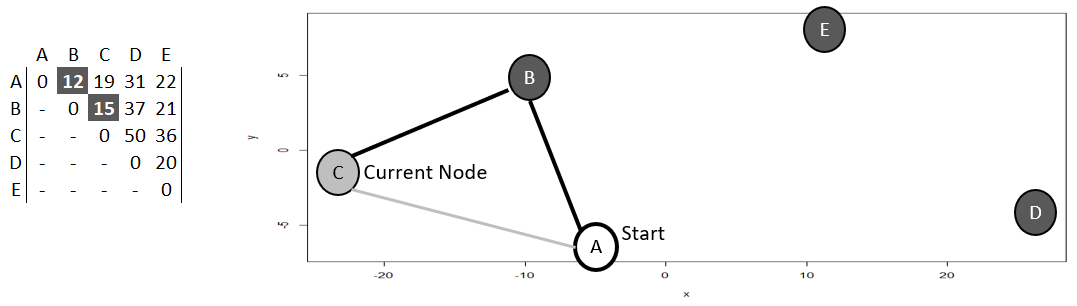}
    \caption{EL subtour identification example - Iteration 1: Arc A-C is added to tour}
    \label{elsub1}
\end{figure}

\begin{figure}[h!]
	\centering
    \includegraphics[width=12.5cm]{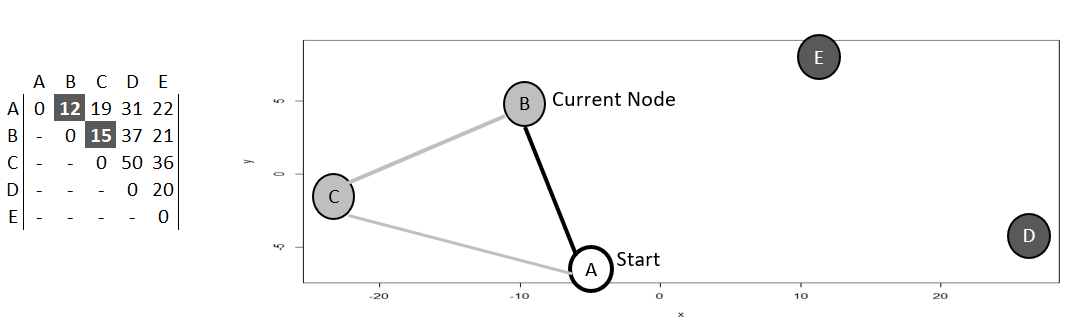}
    \caption{EL subtour identification example - Iteration 2: EL finds node B connected to node C}
    \label{elsub2}
\end{figure}

\begin{figure}[h!]
	\centering
    \includegraphics[width=12.5cm]{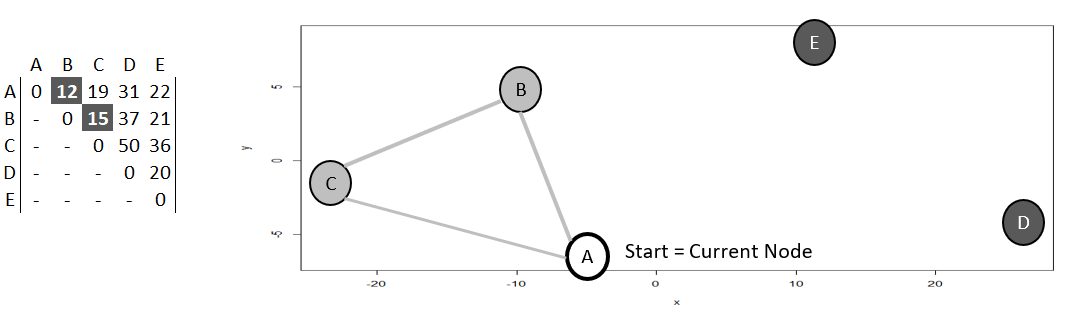}
    \caption{EL subtour identification example - Iteration 3: EL finds start node A connected to node B.  Since Counter is below |N| = 5 EL identifies arc A-C resulted in a subtour}
    \label{elsub3}
\end{figure}

 When applied to the earlier example, after adding edge A-C the heuristic identifies node A as the starting node and Node C as the current node (seen in Figure \ref{elsub1}).
The heuristic finds Node C has a degree of 2 and finds the other connected arc C-B. Node B becomes the current node and the heuristic verifies that the current node is not the same as the start node (seen in Figure \ref{elsub2}).
Once again, the heuristic finds the new current node, Node B, has a degree of 2, finds the other connected arc B-A, and updates the current node to Node A. This time when the heuristic checks the current and start node, it realizes they are the same (Figure \ref{elsub3}). Since the counter is less than \textit{N}, the heuristic marks that a subtour has formed and that arc C-A is not valid.
Pseudocode for the Non-Directional variant of this methodology can be found in Algorithm \ref{fig:ExhloopPseudo}. The EL can also be modified to handle a directional methodology by splitting the `Degree' array into a `To' and `From' array and rather than checking if current node has a degree of 2, check if the current node has a value of 1 in 'From' array

\begin{algorithm}
\caption{Exhaustive Loop (Non-Directional) Pseudocode}
\label{fig:ExhloopPseudo}
\begin{algorithmic}[1]
\State Initialize Variables
\State Sort edges: Shortest to Longest
\While {\textit{Nodes.Visited} \textless Size-1}
\If {Both nodes of current edge have degree \textless 2} 
\State Set Start = Tail of current edge
\State Set Current = Head of current edge
\While {Continue = True}
\If {Current is Tail to Another Edge}
\State Set Next Node = Head of found edge
\If {Next Node = Start}
\State Subtour Formed | Remove Edge
\State Continue = False
\Else 
\State Current = Next
\EndIf
\Else
\State Continue = False
\State Set edge as part of tour
\State Nodes.Visited = Nodes.Visited + 1
\EndIf
\EndWhile
\EndIf
\State Next Edge in List
\EndWhile
\State Connect Hamilton Path
\end{algorithmic}
\end{algorithm}

\subsubsection{Greedy Tracker}
We now introduce a novel way to track the progress of the arc-greedy construction heuristic, and ensure subtours are not created. This new method is called the "greedy tracker" (GT). The GT tracks a node's connection with other nodes when constructing a TSP tour. While we define both non-directional and directional GT variants, it is conceptually easier to visualize the GT using its directional variant on a symmetric instance and then generalize the process for use on asymmetric instances or to the non-directional variant. Thus, the following introduction to the GT utilizes the directional variant on a symmetric matrix and is accomplished using the following structures:
\begin{itemize}
\item $X=$ binary $n$ by $n$ matrix of $x_{ij}$,
\item $F=$ binary $n$ by $1$ array of $f_i$,
\item $T=$ binary $n$ by $1$ array of $t_i$,
\item $x_{ij}=$ 0 if arc from $i$ to $j$ is eligible, greater than 0 if not eligible,
\item $f_i=$ binary for whether node \textit{i} has been left,
\item $t_i=$ binary for whether node \textit{i} has been entered.
\end{itemize}
These structures track each move to prevent sub tours. Given our prior example the initial condition of GT, and associated structures, can be seen in Figure \ref{fig:figure3}:

\begin{figure}[h!]
	\centering
    \includegraphics[width=10cm]{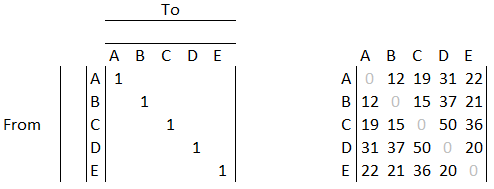}
    \caption{GT Example -  Initialization. X matrix, To and From arrays on left, distance matrix on right for ease of reference. "1s" on diagonal of X matrix indicate illegal arcs.}
  	\label{fig:figure3}
\end{figure}

\par
\begin{figure}[h!]
	\centering
    \includegraphics[width=10cm]{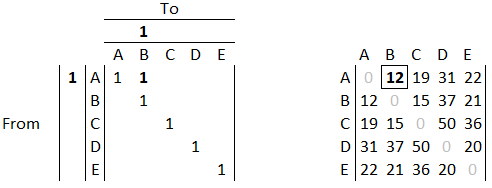}
    \caption{GT Example - Iteration 1, step a. Bold 1s in To and From array, and in X matrix represent the addition of arc A-B to tour, also seen with framed 12 in distance matrix.}
  	\label{fig:figure4}
\end{figure}

 The X (identity), F (From), and T (To) structures can be seen above on the left and, for ease of reference, the associated distance matrix from the TSP instance can be seen on the right. The 1s loaded on the diagonal of the X matrix (where \textit{i=j}) signal ineligible moves. Note that the diagonal on the distance matrix has been colored grey, and set to zero, to correspondingly show these ineligible arcs. The distance matrix indicates that the shortest arc is either from A to B or vice versa, thus arc A to B is selected. The X, F, and T matrices are updated with 1s to indicate this move, as shown in Figure \ref{fig:figure4}.

\par
\begin{figure}[H]
	\centering
    \includegraphics[width=10cm]{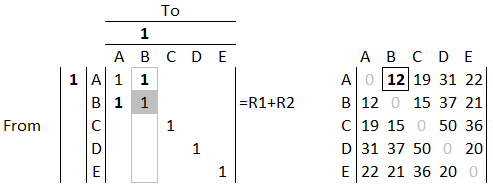}
    \caption{GT Example - Iteration 1, step b. Addition of arc A-B propagates additional 1s in X matrix to capture illegal arcs.}
  	\label{fig:figure5}
\end{figure}

 Next, the column of the X matrix associated with the new arc is processed. Every row where a 1 appears is combined with the 'From' row of the created arc. Figure \ref{fig:figure5} illustrates this operation. As seen in Figure \ref{fig:figure5}, since Row 2 has a 1 in the same column as our new arc, the two rows were combined so that any 1s that were in the Row 1 are now also in Row 2. Note that for the example we only show values of 1 so as to not detract from their purpose of referring to an ineligible move. However, in the code the values in each row will actually be added and values of greater than 1 will appear.
\FloatBarrier \vspace{-.15 in}

\begin{figure}[h!]
	\centering
    \includegraphics[width=10cm]{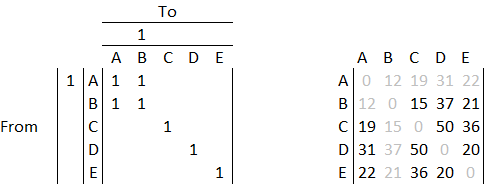}
    \caption{GT Example - Iteration 1, step b. The illegal arcs can be shown for ease of reference as greyed out numbers in distance matrix.}
  	\label{fig:figure6}
\end{figure}

 For ease of reference in this example ineligible values in the distance matrix are turned grey. As can be seen in Figure \ref{fig:figure6}, distances that correspond with a 1 in the X matrix are ineligible moves. Note that any row or column that has a 1 in the T or F arrays is also marked as an ineligible move. This information is utilized in the first step of the next iteration where the shortest available arc is identified.

\begin{figure}[H]
	\centering
    \includegraphics[width=10cm]{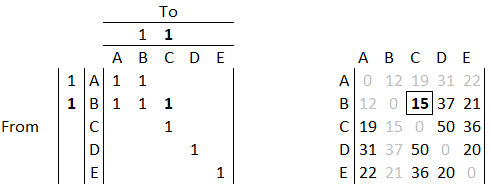}
    \caption{GT Example - Iteration 2, step a. Arc B-C is added to tour, the To array, From array and X matrix are updated.}
  	\label{GTSubtour5}
\end{figure}

 As seen in Figure \ref{GTSubtour5}, the shortest available arc is B-C and once again the X matrix, From, and To arrays are updated with 1s to indicate the move.
\FloatBarrier

\begin{figure}[h!]
	\centering
    \includegraphics[width=10cm]{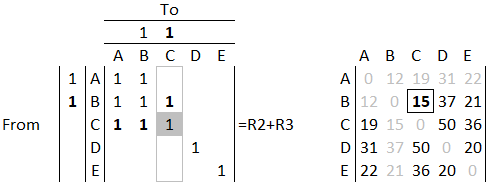}
    \caption{GT Example - Iteration 2, step b. Additional 1s are propagated thru X matrix to prevent usage of illegal arcs.}
  	\label{GTSubtour6}
\end{figure}

The column of X associated with the "To" node of the new arc is processed and every row where a 1 appears is combined with the ``From" row of the created arc which can be seen in Figure \ref{GTSubtour6}.

\begin{figure}[h!]
	\centering
    \includegraphics[width=10cm]{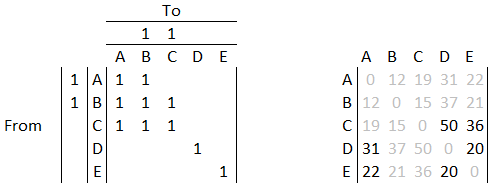}
    \caption{GT Example - Iteration 2, step b. For ease of reference illegal arcs in distance matrix are greyed out.}
  	\label{GTSubtour7}
\end{figure}

All the distances that correspond with a 1 in the X matrix are marked as ineligible moves in the distance matrix, as well as any distances associated with a 1 in the T and F arrays. The resulting step can be seen in Figure \ref{GTSubtour7}. 

The grey numbers in the distance matrix indicates that adding arc A-C is no longer possible because node C already has an edge entering it. This process prevents the formation of the subtour. The process shown above continues until all nodes have been visited which creates a Hamiltonian Path. The final connection to complete the tour is made using the T and F arrays as each will have one index that is still empty. 
\par
\begin{figure}[h!]
	\centering
    \includegraphics[width=9cm]{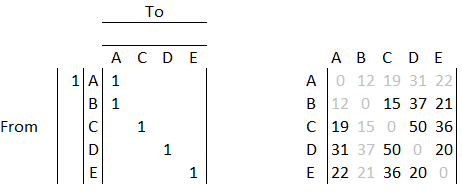}
    \caption{GT Column Delete 1}
  	\label{rowdel1}
\end{figure}

 Real-world implementation of GT requires adjustments to achieve a more linear computational growth for larger instances by reducing the total number of operations that occur within each iteration. This is achieved by removing the addition of values with respect to nodes that have been exited and entered. This decreases the dimensionality of the GT as the tour is constructed and is possible because once a node has been entered (resp. exited) no more arcs may enter (resp. exit) that node. Therefore, it is unnecessary to track what arcs could produce a subtour by entering (resp. exiting) that node. Consider for example the same 5-by-5 instance. After completing the row additions after adding arc A-B, column B can be deleted. Figure \ref{rowdel1} shows the resulting GT and distance matrix.
This process can also be applied to rows to further reduce dimensionality. When working with a non-directional instance a column and row would be deleted after the node had a degree of 2.
\par  Breaking down the process of the row and column delete methodology leads to the realization that fundamentally after an arc is added the GT is only identifying the arc(s) which would form a subtour for the fragment including that new arc.  Bentleys MF on the other hand tracks the tails of every fragment and prevents the head and tail of a fragment from connecting.  There appears to be some advantage in GTs approach as only one number is added to the X matrix at each step. However, the GT must search for and identify the exact arc that needs to be moved, whereas while MF references more nodes and tails, there is no searching necessary. 

Pseudocode for the GT subtour elimination method is found in Algorithm \ref{GTPseudo}.
\begin{algorithm}[h!]
\caption{Greedy Tracker modified Pseudocode}
\label{GTPseudo}
\begin{algorithmic}[1]
\State Initialize Variables
\State Sort edges(i,j): Shortest to Longest
\While {Nodes.Visited \textless Size-1}
\If {To[j]=0 \& XMatrix[i,j]=0}
\State		Nodes.Visited + 1
\State		XMatrix[i,j]=1
\State		From[i]=1
\State		To[j]=1
\State		Row = Intersect(which(X[,j]==0,which(From==0))
\State		Column = Intersect(which(X[,j]==0,which(From==0))
\State		XMatrix[Row,Column]=1
\State	Next Edge in List
\EndIf
\EndWhile
\State Connect Hamilton Path
\end{algorithmic}
\end{algorithm}

\section{Results and Analysis} 
To compare the run times for the MF, EL, and GT subtour elimination methods all three methods were run across 14 symmetric and 9 asymmetric TSP instances.

\subsection{TSP Instances}
TSP instances are available in an online library, TSPLIB, maintained by Ruprecht-Karls-Universitat Heidelberg located in Baden-Wurttemberg, Germany \citep{TSPlib}. For the purposes of this research testing was performed on the instances seen in Table \ref{fig:TSPinstance}, where the alpha prefix is an identifier and the numerical suffix indicates the instance size (in number of nodes).
\begin{table}[h!]
\centering
\caption{TSP Instances used in current testing}
\begin{tabular}{|| c c ||} 
 \hline
 \multicolumn{2}{||c||}{TSP Instances} \\
 \hline
 Symmetric & Asymmetric \\
 \hline
 bays29 & br17  \\
 gr48 & ry48p \\
 gr51 & ft53  \\
 berlin52 & ft70   \\
 pr76 & kro124p  \\
 kroa100 & rgb323   \\
 gr120 & rgb358 \\
 gr130 & rgb403   \\
 gr195 & rgb443  \\
 ts225 & \\
 pma343 & \\
 pcb442 & \\
 dsj1000 & \\
 pr1002 & \\
 \hline
\end{tabular}
\label{fig:TSPinstance}
\end{table}
\subsection{Testing}
Initial tests verified that each subtour method (MF, EL, and GT) resulted in the same tour for all TSP instances. These tests were conducted with both directional and non-directional versions of codes on symmetric TSP instances. In addition, the directional code versions were run on asymmetric TSP instances. 
\par Once testing verified each method produced identical greedy tours; that is all directional code variants produced identical tours, and all non-directional variants produced identical tours, the remaining testing focused on computational run-time comparisons. Each methodology was placed in the same arc-greedy heuristic shell so that testing would fairly compare the speed of the three subtour tracking and elimination methodologies. Bentley \cite{bentley1992fast} and Wang \cite{wang2018distance} each utilized advanced computer techniques (k-d trees) and additional data structures to speed up the process of finding the next shortest arc available. However, since neither of these effect the speed of the subtour tracking and elimination methodologies they were not utilized. 
\par Speed tests were conducted utilizing the R package ``microbenchmark."  100 iterations of each code were run to create summary statistics on 13 different symmetric TSP instances and 9 asymmetric instances. Both symmetric and asymmetric instances were tested to determine if symmetry effected run time.

\subsection{Symmetric Instance Results}
Mean run times for a variety of symmetric TSP instances utilizing each of the methodologies can be seen in Table \ref{SpeedTable1}.  
\begin{table}[h!]
\centering
\caption{Greedy subtour Methodology Run Times (Symmetric Instances)}
\begin{tabular}{||l | c c c | c c c ||} 
 \hline
 Milliseconds & \multicolumn{3}{c|}{Directional} &  \multicolumn{3}{c||}{Non-Directional} \\
 \hline
 Instance & EL & MF & GT & EL & MF & GT \\
 \hline
 bays29 & 23.2 & 52.0 & 22.4 & 29.7 & 44.2 & 24.4 \\
 gr48 & 25.4 & 56.1 & 25.0 & 30.5 & 46.8 & 25.8 \\
 eil51 & 26.7 & 53.2 & 24.4 & 32.3 & 50.4 &28.1 \\
 berlin52 & 26.6 & 53.1 & 24.1 & 31.4 & 47.9 &26.3 \\
 pr76 & 30.8 & 56.9 & 28.5 & 32.8 & 46.9 & 29.9 \\
 kroa100 & 36.8 & 60.8 & 34.5 & 36.5 & 50.4 & 30.8 \\
 gr120 & 41.9 & 66.5 & 43.1 & 40.5 & 52.2 & 34.9 \\
 ch130 & 48.9 & 70.4 & 43.7 & 43.3 & 52.9 & 35.7 \\
 rat195 & 76.8 & 98.6 & 73.6 & 54.5 & 65.0 & 50.3 \\
 ts225 & 97.8 & 113.9 & 95.4 & 61.8 & 68.1 & 57.3 \\
 pma343 & 193.7 & 193.9 & 177.0 & 121.2 & 111.9 & 107.7 \\
 pcb442 & 363.7 & 312.8 & 317.2 & 177.0 & 166.1 & 180.5 \\
 dsj1000 & 1.667(s) & 1.341(s)& 1.440(s) & 906.9 & 660.1 & 750.2 \\ 
 pr1002 & 1.595(s) & 1.308(s) & 1.374(s) & 863.3 & 637.3 & 761.8 \\
 \hline
\end{tabular}
\label{SpeedTable1}
\end{table}
When looking at the directional variants, the GT tends to be the fastest methodology on small instances followed by EL and MF. Once instance size reaches around 442, MF takes over as the fastest method for eliminating subtours. This largely is due to it's linear growth in operation count as instance size grows. For larger instances, the heuristic is conducting the same number of operations at each step. While the operations are slower for small instances, once the problem becomes larger it proves to be the most efficient. 
\par As expected all Non-Directional subtour elimination variants run faster on symmetric instances versus their Directional counterparts.  GT remains the fastest for instances up to around 442 nodes in size. EL is second fastest up till 343 nodes. At 442 nodes MF is fastest method.  

\ref{SpeedTable1}.

\subsection{Asymmetric Instance Results}
\par The directional variants of each subtour elimination codes were also run on Asymmetric TSP instances to compare runtimes to determine if any trends changed. The mean runtimes are in Table \ref{SpeedTable2}.
\begin{table}[h!]
\centering
\caption{Greedy subtour Methodology Run Times (Asymmetric Instances)}
\begin{tabular}{||l | c c c ||} 
 \hline
 Milliseconds & \multicolumn{3}{c|}{Directional} \\
 \hline
 Instance & EL & MF & GT \\
 \hline
 br17 & 22.2 & 52.5 & 21.1  \\
 ry48p & 24.5 & 53.9 & 23.2 \\
 ft53 & 25.6 & 54.0 & 24.7  \\
 ft70 & 28.5 & 56.5 & 27.4  \\
 kro124p & 35.9 & 64.4 & 34.6  \\
 rgb323 & 230.3 & 195.5 & 199.1  \\
 rgb358 & 242.7 & 219.7 & 231.3  \\
 rgb403 & 311.9 & 275.2 & 274.5  \\
 rgb443 & 366.5 & 315.3 & 330.5  \\
 \hline
\end{tabular}
\label{SpeedTable2}
\end{table}
\par For the asymmetrical instances we note that MF became the fastest methodology when the instance size reached 358 nodes, but was beaten by the GT at an instance size of 403 nodes. This variation leads us to believe that some of the subtour methodologies may have computational advantages for specific TSP instances dependent on how the tour is constructed. For example, specific instances may have nodes spaced in such a fashion that for a majority of the tour construction the tracking methodology is maintaining a small number of large fragments. It is possible that one of the methodologies is computationally more efficient for these instances and may be computationally slower in TSP instances were nodes force an arc-greedy tour construction that results in many small fragments to be tracked. Future analysis of instance geometry and resulting runtimes should be conducted to test whether this hypothesis holds any merit. With the exception of the variation seen for the 403 node instance, prior overall trends from the Symmetric instances remain, where GT is competitive for small to medium sized asymmetric instances, but MF is fastest for larger instances.

\subsection{Future Improvements/Analysis}
The portion of the GT most susceptible to computational growth is the search to identify what tail of the path to move. If this search process growth can be limited, it is possible that the modified GT could outperform MF for larger instances as well. Some possible methodologies to limit computational growth include a better implementation of the row and column delete methodology in conjunction with a new row and column generation methodology. Size of the search operations could also be reduced drastically especially during the early iterations of the arc-greedy heuristic by only generating nodes and tails as needed.
\par The significance of separating the subtour tracking methodologies from the underlying greedy heuristic methodology is the ability to then use these subtour elimination methodologies to develop new fragment heuristics to build viable TSP tours. For example consider the constructive heuristic called the Ordered-Greedy (OG) heuristic. The OG heuristic is a node-greedy heuristic that takes as input a complete ordered list of nodes. Starting at the top of the list, each node is considered in turn and the available set of choices $S_j$ at each step is limited to the feasible arcs originating at that node. What differentiates the OG from NN, another node-greedy heuristic, is that multiple fragments may exist during the tour construction.
\par The motivation for the OG heuristic is to apply a more structured approach in which nodes are given priority in connecting to their nearest neighbors. Nodes higher in the list have maximum flexibility with minimal concern for node degree or subtours and thus typically choose better arcs than nodes later in the list which experience significantly less degrees of freedom in their legal arc choices. The quality of the solution found is thus heavily dependent on the order of the list. 
\par
\begin{figure}[H]
	\centering
    \includegraphics[width=10cm]{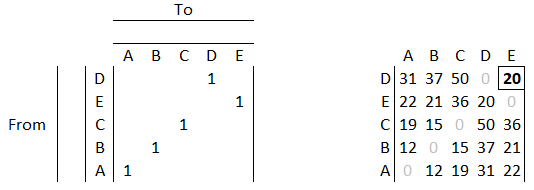}
    \caption{OG example - Iteration 1.  OG picks greediest legal arc incident to node D.}
  	\label{fig:OG1}
\end{figure}

\begin{figure}[H]
	\centering
    \includegraphics[width=10cm]{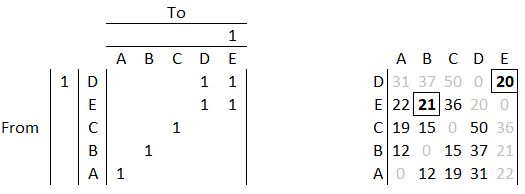}
    \caption{OG example - Iteration 2.  OG picks greediest legal arc incident to node E}
  	\label{fig:OG2}
\end{figure}

\begin{figure}[h!]
	\centering
    \includegraphics[width=10cm]{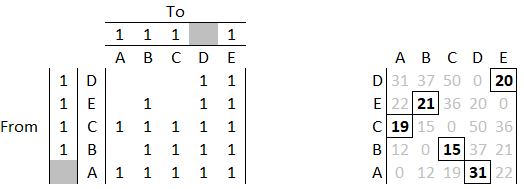}
    \caption{Ordered-Greedy Step 3. X matrix on the left, reordered distance matrix on the right.}
  	\label{fig:OG3}
\end{figure}

 To introduce the methodology of the OG heuristic, consider the following example. In this example an ordered list of D,E,C,B,A has been, through some unspecified fashion, predetermined. This ordering of the nodes list is reflected, for ease of reference, in the X matrix and To and From arrays on the left-side, and in the distance matrix on the right-side of Figure \ref{fig:OG1} whose rows are now sorted according to this list order. The constructive heuristic now makes greedy decisions starting at the top of this list and working down. The first greedy decision is made with respect to node D. 
\FloatBarrier \vspace{-.15in}
\vspace{-.15in}
The greediest, or shortest arc, from node D is arc D-E as indicated above. This arc and its associated node is tracked via the GT so the next decision can be made. The next decision is made with respect to node E. This is not due to node E being the head of the previous arc added, but rather because it is second in the provided ordered list: D,E,C,B,A. Looking at the row in the distance matrix associated with node E along with the GT output that captures ineligible moves (as seen in Figure \ref{fig:OG2}), it can be seen that the shortest legal arc available is arc E-B.
\FloatBarrier
This process continues row by row until the final row is reached which is where the To and From arrays are scanned to find the final legal arc as seen in Figure \ref{fig:OG3}.
\FloatBarrier
After adding arc A-D, the resulting tour becomes A-D-E-B-C-A which is also the optimal tour for this TSP instance.  Pseudocode for the OG is found in Algorithm \ref{fig:OGPseudo}.
\begin{algorithm}[h!]
\caption{Ordered-Greedy Pseudocode}
\label{fig:OGPseudo}
\begin{algorithmic}[1]
\State Initialize Variables
\State Generate Order
\State Nodes.Visited = 0
\While {Nodes.Visited \textless Size-1}
\State Moves = arcs leaving Order[Nodes.Visited+1]
\State Moves[To==True]= Inf
\State Moves[XMatrix[Order[Nodes.Visited+1],]]= Inf
\State minmove = min(Moves)
\State Get First index i of Moves where Moves[i]=minmove
\State Add Arc(Order[Nodes.vistied+1],i) to tour
\State Track moves with \textbf{Greedy Tracker}
\State Nodes.Visited = Nodes.Visited+1
\EndWhile
\State Connect Hamilton Path
\end{algorithmic}
\end{algorithm}

\section{Conclusion}
As an NP-hard combinatorial optimization problem, the TSP is often solved via heuristic methodologies. One of the biggest considerations when constructing solutions is avoiding subtours, or a loop of interconnected nodes that prevents a single continuous tour amongst all cities within the instance. This paper introduced a novel subtour elimination methodology for the arc-greedy heuristic that is compared to two known subtour elimination methodologies. Computational results were generated across multiple TSP instances for each method.
\par When utilizing an arc-greedy type heuristic, additional steps must be taken to ensure that subtours are avoided and resulting tour is a valid TSP solution. This paper recognized two accepted arc-greedy subtour elimination methodologies, the Exhaustive loop and Bentley's MF, and compared them to our Greedy Tracker. The comparison utilized both directional and non-directional variants of each code on 14 symmetric TSP instances and the directional variants on 9 asymmetric instances.
\par The results of the comparison between each of these arc-greedy subtour elimination methodologies showed that the GT was the fastest tracking methodology for small to medium sized instances.  However, Bentley's MF maintains the computational advantage for larger instances.
\par However, these results also indicated that given a more efficient coding implementation of methodology used for the X Matrix, the GT could become the preferred methodology for all instance sizes. For future research, the GT should be modified to handle a new row/column generation and delete technique to minimize the computational time utilized in the searching portions of the GT.



\bibliographystyle{unsrt}
\bibliography{bibli.bib} 

\end{document}